\pdfoutput=1

\documentclass[11pt]{article}

\usepackage[final]{acl}

\usepackage{times}
\usepackage{latexsym}
 \usepackage{multirow}
 \usepackage[many]{tcolorbox} 
\usepackage[T1]{fontenc}

\usepackage[utf8]{inputenc}

\usepackage{microtype}

\usepackage{inconsolata}
\usepackage{booktabs}
\usepackage{graphicx}
\usepackage{amssymb}

\title{Developing multilingual speech synthesis system for Ojibwe, Mi'kmaq, and Maliseet}

\author{
    Shenran Wang \textsuperscript{1} \\
    \And
    Changbing Yang \textsuperscript{1} \\
    \And
    Mike Parkhill \textsuperscript{2} \\
    \And
    Chad Quinn \textsuperscript{3} \\
    \AND
    Christopher Hammerly \textsuperscript{1} \\
    \And
    Jian Zhu \textsuperscript{1} \\
    \AND
    \parbox{\textwidth}{
        \normalfont
        \centering
        \textsuperscript{1}University of British Columbia \space
        \textsuperscript{2}SayITFirst \space
        \textsuperscript{3}CultureFoundry \\
        \texttt{\{shenranw, cyang33\}@mail.ubc.ca, mikepark@sayitfirst.ca, \\
        chadquinn@culturefoundrystudios.com, \{chris.hammerly, jian.zhu\}@ubc.ca}
    }
}

\newtcolorbox{boxK}{
    sharpish corners, 
    boxrule = 0pt,
    toprule = 4.5pt, 
    enhanced,
    fuzzy shadow = {0pt}{-2pt}{-0.5pt}{0.5pt}{black!35} 
}

\begin{document}
\maketitle
\begin{abstract}
We present lightweight flow matching multilingual text-to-speech (TTS) systems for Ojibwe, Mi'kmaq, and Maliseet, three Indigenous languages in North America. Our results show that training a multilingual TTS model on three typologically similar languages can improve the performance over monolingual models, especially when data are scarce. Attention-free architectures are highly competitive with self-attention architecture with higher memory efficiency. Our research not only advances technical development for the revitalization of low-resource languages but also highlights the cultural gap in human evaluation protocols, calling for a more community-centered approach to human evaluation. 
\end{abstract}

\begin{table*}[]
\centering
\small
\begin{tabular}{rrrrrrrrr}
\toprule
\multirow{2}{*}{Language} & \multirow{2}{*}{Speaker} & \multirow{2}{*}{Gender} & \multicolumn{2}{c}{Train} & \multicolumn{2}{c}{Dev} & \multicolumn{2}{c}{Test} \\
             &     &       & Duration     & Samples    & Duration    & Samples   & Duration    & Samples    \\\midrule
 Ojibwe & JJ      & M                  &    11h 49min 23s         &    11,062        &  6min 38s           &    100       &  6min 18s           &   100         \\
 Ojibwe & NJ    & F     &  1h 41min 14s             &      2404        &    4min 21s        &   100          &    4min 9s       &              100        \\
 Mi'kmaq & MJ & F & 2h 22min 57s & 1116 & 12min 22s & 100 & 12min 30s & 100 \\
  Maliseet & AT & M & 7h 15min 25s & 3628 & 12min 16s & 100 & 12min 27s & 100 \\\bottomrule
\end{tabular}
\caption{A summary of the Indigenous speech corpora in this study.}
\label{table:dataset}
\end{table*}

\section{Introduction}
Many world languages are currently endangered, especially those spoken by historically marginalized and Indigenous communities. Language revitalization and reclamation is an ongoing effort to ensure continued language vitality for community self-determination and well-being \cite{oster2014cultural,mccarty2018community,bird-2020-decolonising}. Among recent efforts of language revitalization, TTS technology is valued as a potential tool to assist the education of Indigenous languages, as TTS models can flexibly synthesize diverse learning materials to guide pronunciation learning \cite{pine-etal-2022-requirements,pine2024speech}. 

In general, speech synthesis for Indigenous languages is underdeveloped compared to the majority of languages. The main barrier to developing TTS technologies for Indigenous communities with oral traditions is still the lack of data \cite{pine-etal-2022-requirements,pine2024speech}. There are recent efforts to develop speech synthesis systems for low-resource and Indigenous languages, including Mundari \cite{gumma-etal-2024-muntts}, Kanien’kéha (also known as
Mohawk; Iroquoian), Gitksan (Tsimshianic),
SEN COTEN (Coast Salish) \cite{pine-etal-2022-requirements,pine2024speech}, Plains Cree (Central Algonquian) \cite{harrigan2019preliminary} and Ojibwe \cite{hammerly-etal-2023-text}. Yet there is still room for improvement and development in this space.

In this study, we continue this line of effort and develop TTS systems for \textbf{Ojibwe}, \textbf{Mi'kmaq}, and \textbf{Maliseet}, the latter two of which haven't received any attention from the speech processing community yet. Our study explicitly tackles several challenges in designing speech technology for Indigenous communities. 
\begin{itemize}
    \item First, it is generally impractical to bring Indigenous members to labs for recording, so we demonstrate a community-centered approach to allow speakers to record their own voices at their own pace. 
    \item Secondly, as collecting Indigenous speech at scale is difficult, we show that training a flow matching multilingual TTS models \cite{mehta2024matcha} with typologically similar language varieties can help improve the synthesis performance in low-resource settings. 
    \item Thirdly, since the TTS system is likely to be deployed in common computing devices, we also implemented attention-free architectures, including FNet \cite{lee-thorp-etal-2022-fnet}, Mamba2 \cite{mamba2} and Hydra \cite{hydra} that closely match the performance of self-attention models in TTS but are generally more efficient in deployment. \item Finally, we also discuss the need to adapt current experimental paradigms to better work with Indigenous communities. 
\end{itemize}

The code is available at: \url{https://github.com/ShenranTomWang/TTS}. 

\section{Data Collection}
\paragraph{Languages}
We worked closely with speakers from three Indigenous languages of Canada: \textbf{Ojibwe}, \textbf{Mi'kmaq}, and \textbf{Maliseet}. The three languages are genetically related. Ojibwe is spoken around the Great lakes of North America and is part of the Central branch of the Algonquian family, while Mi'kmaq and Maliseet are spoken in the Maritimes and are classed within the Eastern branch of the Algonquian family. According to estimates from the 2021 Statistics Canada Survey, there are 25,440 speakers of Ojibwe, 9,000 speakers of Mi'kmaq, and 790 speakers of Maliseet \cite{robertson2023census}. All language communities are actively involved in significant efforts to document and ensure the continued vitality of their languages. 

\paragraph{Data collection}
Most Indigenous speakers fluent in their own languages are senior speakers. It is infeasible to bring them to a sound-proof lab for recording at a university. Instead, we adopted a community-centered approach that allows the speakers to have full control over the speech recording process in the comfort of their own home, following the protocol from a prior study \cite{hammerly-etal-2023-text}. 

In each case, we used texts identified by the community members as representative of their dialect and writing system as the basis for the data set. These texts were then split into individual utterances (complete sentences or phrases) and loaded into the prompting and recording program SpeechRecorder \cite{draxler2004speechrecorder}. The program allows speakers to read and record utterances at their own pace, easily re-record in the case of an error or disfluency, and package and upload recorded utterances into secure cloud storage as they complete them. 


\paragraph{Data partition}
We resampled the recorded audio to 22,050Hz. For each speaker, we reserved 100 random samples for validation and another 100 random samples for test. The rest of the speech samples were used for model training. The detailed statistics of our data were summarized in Table~\ref{table:dataset}. Since each of our datasets has a different size, we applied oversampling to our multilingual training dataset by duplicating training samples such that they contain roughly the same duration for each speaker.


\begin{table*}[!ht]
    \centering
    \small
    \resizebox{1\textwidth}{!}{
        \begin{tabular}{lrrrrrrrr}
        \toprule
            \textbf{Model} & \textbf{F0 RMSE↓} & \textbf{LAS RMSE↓} & \textbf{MCD↓} & \textbf{PESQ↑} & \textbf{STOI↑} & \textbf{VUV F1↑} & \textbf{FID↓} & \textbf{MOS↑} \\ \midrule
\textcolor{blue}{\textbf{Ojibwe JJ}} Natural &  - & - & - & -& - & - & - & 4.16 \\
Monolingual Self-Attention & 57.317 & 5.284 & 22.044 & 1.228 & 0.036 & 0.845 & 0.005 & 2.71 \\
Multilingual Mamba2 & \textbf{55.982} & \textbf{4.813} & \textbf{18.414} & 1.229 & 0.035 & \textbf{0.845} & 0.004 & 3.00 \\
Multilingual Hydra & 56.806 & 5.147 & 18.849 & \textbf{1.276} & \textbf{0.036} & 0.842 & 0.004 & 3.25 \\
Multilingual FNet & 58.871 & 5.720 & 19.463 & 1.170 & 0.036 & 0.824 & 0.006 & \textbf{3.42}\\
Multilingual Attention & 56.454 & 4.859 & 18.427 & 1.240 & 0.034 & 0.843 & \textbf{0.004} &  2.67 \\
 \midrule
\textcolor{blue}{\textbf{Ojibwe NJ}} Natural &  - & - & - & -& - & - & - & 4.74 \\ 
Monolingual Self-Attention & \textbf{80.511} & \textbf{6.198} & 17.568 & 1.120 & \textbf{0.033} & 0.825 & 0.006 & 4.67 \\
Multilingual Mamba2 & 89.879 & 6.311 & 17.506 & 1.111 & 0.028 & 0.820 & \textbf{0.005} & 4.56 \\
Multilingual Hydra & 87.509 & 6.676 & 18.036 & 1.128 & 0.029 & 0.830 & 0.006 & 4.70 \\
Multilingual FNet & 97.015 & 6.728 & 19.271 & 1.099 & 0.030 & 0.787 & 0.012 & 4.69\\
Multilingual Attention & 86.446 & 6.462 & \textbf{17.414} & \textbf{1.147} & 0.028 & \textbf{0.835} & 0.006 & \textbf{4.77} \\
\midrule
\textcolor{blue}{\textbf{Mi'kmaq MJ}} Natural &  - & - & - & -& - & - & - & - \\ 
Monolingual Self-Attention  & 138.890 & 8.614 & 21.720 & 1.110 & \textbf{0.039} & 0.640 & 0.006 & - \\
Multilingual Mamba2 & 139.574 & 7.831 & 22.060 & 1.165 & 0.038 & 0.643 & 0.005 & - \\
Multilingual Hydra & \textbf{138.157} & \textbf{7.128} & 21.694 & \textbf{1.210} & 0.038 & 0.649 & 0.005 & - \\
Multilingual FNet & 144.566 & 7.761 & 21.748 & 1.183 & 0.038 & 0.631 & 0.005 & - \\
Multilingual Attention & 138.365 & 7.357 & \textbf{21.588} & 1.165 & 0.037 & \textbf{0.667} & \textbf{0.003} & - \\
 \midrule
 \textcolor{blue}{\textbf{Maliseet AT}} Natural &  - & - & - & -& - & - & - & - \\ 
Monolingual Self-Attention  & 79.807 & 9.066 & 19.576 & 1.262 & 0.031 & 0.657 & 0.005 & - \\
Multilingual Mamba2 & 77.725 & 8.565 & 19.152 & 1.213 & \textbf{0.038} & 0.727 & 0.007 & - \\
Multilingual Hydra & 79.834 & 8.414 & \textbf{18.129} & \textbf{1.500} & 0.035 & 0.728 & 0.006 & - \\
Multilingual FNet & 76.308 & 8.947 & 19.058 & 1.259 & 0.037 & 0.723 & 0.008 & - \\
Multilingual Attention & \textbf{75.267} & \textbf{8.032} & 18.173 & 1.316 & 0.032 & \textbf{0.742} & \textbf{0.005} & - \\
\hline
  \end{tabular}
    }
    \caption{Evaluation results for each speaker across all models in \texttt{float32}.}
    \label{table:all_results}
\end{table*}

\section{Method}
\subsection{MatchaTTS}
Our system is built upon Matcha-TTS \cite{mehta2024matcha}, a fast TTS model based on conditional flow matching, a class of probabilistic generative model capable of generating high-fidelity image and audio \cite{lipman2023flow}. The original Matcha-TTS consists of a text encoder, a duration predictor, and a flow matching decoder. The text encoder transforms the text input into hidden states, which are then upsampled to the output length based on the duration predictor. The flow matching decoder predicts the final mel spectrogram through iterative denoising steps conditioning on the upsampled hidden states. 

The original MatchaTTS model was only designed for single-speaker TTS. For multilingual speech synthesis, we added learnable speaker and language embeddings for each unique speaker and language, a common technique for multilingual models \cite{cho22_interspeech}. Both embeddings were concatenated with the output of the text encoder, which was then fed into the flow-matching decoder for mel-spectrogram prediction. By default, the flow-matching decoder uses 10 inference steps to perform inference. 

\subsection{Sequence mixing layers}
The multilingual MatchaTTS utilizes attention for sequence mixing with 40M parameters, yet its quadratic complexity is not ideal for efficient deployment. Here we also explore different attention-free layers that can also mix information across sequences to improve the efficiency of MatchaTTS. We replace self-attention with each of the following layers. For cross-attention, we concatenate the inputs and put them through each layer.

\paragraph{Mamba2}
Mamba2 \cite{mamba2} is a selective state-space model (SSM)\cite{guefficiently,gu2023mamba} that can perform sequence mixing with subquadratic complexity. SSMs have been shown to be effective in speech generation tasks \cite{zhang2024mamba,miyazaki24_interspeech}. 
In Mamba2, the selective SSM can be formulated as follows:
$$h_t = \mathbf{\overline{A}}_th_{t - 1} + \mathbf{\overline{B}}_tx_t$$
$$y_t = \mathbf{C}_th_{t}$$
where $\mathbf{\overline{B}_t}$ and $\mathbf{C_t}$ are input-dependent weights and $\mathbf{\overline{A}_t}=\alpha_t\mathbf{I}$ is a diagonal matrix. The input-dependent weights allow Mamba2 to selectively focus on the information across time steps, making it effective for sequence processing. 
Mamba2 is closely related to transformers. If $\mathbf{\overline{A}_t}=\mathbf{I}$, it is equivalent to the formulation of linear attention \cite{katharopoulos2020transformers,mamba2}.

In our TTS model, we replaced the attention modules of MatchaTTS with Mamba2 blocks. Noticeably, Mamba2 modules have more parameters than attention modules. In order to keep the total number of parameters consistent, we shrunk the encoder and decoder hidden dimension size by $\frac{3}{4}$, resulting in around 38M parameters in total.

\paragraph{Hydra} As the original Mamba2 is uni-directional, Hydra \cite{hydra} is a bidirectional extension of Mamba2 but still maintains the subquadratic complexity. Below we provide an overview of Hydra.

State-space models, as discussed before, can be formulated by:
$$y = \texttt{SSM}(\mathbf{\overline{A}}, \mathbf{\overline{B}}, \mathbf{C})(x) = \mathbf{M}x$$
Where $x$ is the input, $y$ is the output. Our goal is then to find the matrix $\mathbf{M}$ with desired properties. Current SSMs such as Mamba2 use semiseparable matrices for $\mathbf{M}$. Hydra takes a step further and uses quasiseparable matrices for $\mathbf{M}$, whose computation complexity remains subquadratic and has the nice properties of being able to process inputs in order and reverse. Formally, a matrix is N-quaiseparable iff any submatrix from either the strictly upper or lower triangle has a rank of at most N. Specifically, quasiseparable matrices can be decomposed into semi-separable matrices via:
\[
    \scalebox{0.7}{
        $QS(x) = \texttt{shift}(SS(x)) + \texttt{flip}(\texttt{shift}(SS(\texttt{flip}(x)))) + \mathbf{D}x$
    }
\]
Where $QS(\cdot)$ and $SS(\cdot)$ denote matrix mulplications of quasiseparable and semiseparable matrices respectively, $\texttt{flip}(\cdot)$ denotes the action of reversing the input, $\texttt{shift}(\cdot)$ refers to the action of shifting the input one position to the right (padding with 0 at the beginning), and $\mathbf{D}$ is a diagonal matrix. The $SS()$ operation allows for the selection of any SSMs and we selected the selective SSM in Mamba2. This allows Hydra to perform bidirectional sequence mixing in linear complexity.

While Hydra has not been applied to TTS yet, its bidirectionality makes it potentially more powerful than Mamba2. Hydra layers were used to replace all attention modules in MatchaTTS. Hydra also has more parameters than attention, therefore we also shrunk the encoder and decoder hidden dimension size by $\frac{3}{4}$, resulting in around 39M parameters in total.

\paragraph{Discrete Fourier Transform} Discrete Fourier Transform has proven to be a viable sequence mixing method with a complexity of $O(L\log L)$\cite{lee-thorp-etal-2022-fnet} and works well for speech \cite{chen2024train}. We replaced all attention modules of the MatchaTTS with the FFT layer in FNet.

The FFT layer performs a 2D Fast Fourier Transform, on hidden dimensions and on the sequence dimension of the input and eventually takes the real part of the output. Formally, it can be formulated as:
$$y = \mathbb{R}(\mathcal{F}_{seq}(\mathcal{F}_{h}(x)))$$
Here, $\mathbb{R}(\cdot)$ denotes the action of obtaining real parts of the input, and $\mathcal{F}_{dim}(\cdot)$ denotes the action of performing FFT on the \texttt{dim} dimension of input.

By the duality of the Fourier transform, FNet can be thought of as alternating between multiplications and convolutions. Since this operation is parameter-free, the FNet model has only around 31M parameters.

\begin{table*}[]
\centering
\small
\begin{tabular}{lrrrrr}\toprule
                          & Batch Size      & \textbf{Self-attention} & \textbf{Mamba2} & \textbf{Hydra} & \textbf{FNet} \\\midrule
Throughput (generated speech/s) & 400 &    273.83            &   245.54     &  198.99     &   241.05   \\
Real-time factor & 1       &  0.03         &  0.06      &  0.06     & 0.03     \\
Memory usage & 400 & 4.6G           & 2.3G   & 2.4G  & 2.5G \\
Memory usage & 1  & 245M           & 202M   & 235M  & 230M \\\bottomrule
\end{tabular}
\caption{The time and memory efficiency of different sequence-mixing layers in \texttt{float32} on a single A100 40G.}
\label{tab: efficiency}
\end{table*}

\section{Experiments}
\paragraph{Training} As these languages all use a phonetically transparent Latin alphabet, we used a simple character-based tokenizer to tokenize all sentences. Punctuations were all removed except for the apostrophe in Ojibwe, which plays a role in Ojibwe phonology. Monolingual models were trained for each individual speaker, whereas multilingual models were trained on all speakers with different sequence mixing layers. All experiments were run on a single A100 40GB GPU for a fixed 200 epochs. Full training details are available in Appendix~\ref{app:training_details}.

\paragraph{Vocoder} For waveform generation, we trained a Vocos vocoder \cite{siuzdak2024vocos} on all training samples. Vocos is a frequency domain vocoder that closely matches the performance of time-domain vocoders like Hifi-GAN \cite{kong2020hifi} and diffusion-based vocoder like Fregrad \cite{nguyen2024fregrad} but with much higher throughput. Since vocoder is not the focus, we provided their evaluation results in Appendix~\ref{app:vocoder}.

\section{Results and Discussions}

\paragraph{Objective Evaluation} We perform our objective evaluation results with Fundamental Frequency Root Mean Square Error (F0 RMSE), Log-amplitude RMSE (LAS RMSE), Mel Cepstral Distortion (MCD), Perceptual Evaluation of Speech Quality (PESQ), Short-Time Objective Intelligibility (STOI), Voiced/Unvoiced F1 (VUV F1) and MFCC Frechet Distance (FID), similar to contemprary works \cite{li-etal-2024-cm,lv2024freevfreelunchvocoders}. 

Results in Table~\ref{table:all_results} suggest that multilingual models generally outperform monolingual models in all languages. Training on typologically similar languages does help alleviate the lack of data for individual languages, since the model can learn from the commonalities in these languages. Such findings can also provide guidance for the future collection of Indigenous speech datasets. We can prioritize dataset diversity over quantity, as a large quantity of speech data from a single language is also hard to collect.

While the self-attention MatchaTTS dominates most objective metrics, other attention-free architectures also match its performance closely. No single model dominates all objective metrics. Hydra's performance is particularly close to self-attention, suggesting that it is a strong competitor. Its bidirectional nature also allows it to outperform Mamba2. FNet underperforms all other models due to its parameter-free nature.

In terms of computational efficiency, as shown in Table~\ref{tab: efficiency}, all attention-free architectures are much more memory-efficient than self-attention models, and memory saving is more prominent when the batch size is large. However, the attention-free architectures do not necessarily reduce computation time, presumably because our model is small enough that their advantages are not obvious. 

\paragraph{Subjective Evaluation}
Despite these challenges, in evaluating the current work, we designed separate mean opinion score (MOS) surveys for each language. For each TTS voice, the survey included 10 generated utterances from each of the five models and 10 utterances of natural speech. The detailed design is described in Appendix~\ref{subjective}. We were able to recruit three raters for Ojibwe but one did not complete the survey. For Mi'kmaq and Maliseet, we were not able to obtain MOS rating due to the limited number of speakers. Generally speaking, the MOS ratings are largely consistent with the objective metrics (see Table~\ref{table:all_results}).

As recently discussed in \citet{pine2024speech}, there are many challenges and questions to be raised when conducting a subjective evaluation of speech synthesis with Indigenous communities. We also find that, due to the gap in cultural norms, the use of standard measures like MOS and the current experimental paradigm may not always be viable in determining the quality of synthetic speech. For example, despite our instructions, one Ojibwe rater rated 5 for all Ojibwe NJ's voices, regardless of whether it was natural or synthetic. We believe this may have been due to a reluctance to comment negatively on the voice, even when it was synthetic. The concept of participating in controlled experiments and judging synthetic voices, in general, is not a natural task, and cultural norms can amplify this. This implies that researchers working with Indigenous communities should design more creative measures that also conform to the cultural norms of the relevant community. We plan to conduct such work as we continue development of these systems

\section{Conclusion}
In this paper, we report our ongoing efforts to develop TTS systems with and for the Indigenous community. Our experiments demonstrate that training multilingual TTS models on similar languages can partially compensate for the lack of data for individual languages. In the future, we will be working with the relevant communities and schools to deploy these systems for Indigenous language education. 

\section{Ethical statements}
Our research would not have been possible without the support of the Indigenous communities involved. The subjective evaluation experiments were approved by the institutional ethics review committee. All Indigenous participants in the study, including the voice donors and raters, participated voluntarily and received fair compensation for their contributions. 

The goal of our research is to develop TTS tools for Indigenous communities. We are currently actively working with learners and teachers learning these Indigenous languages at school. However, TTS technology might potentially be misused for impersonation and deception, which can be particularly dangerous for the Indigenous communities as they are not frequently exposed to such technologies. We will continue to work alongside these communities to inform them about the benefits as well as security concerns of speech technologies.  

\section{Limitation}
Our study is still limited in several aspects. First, as all speech recordings were recorded at the speakers' own residence, there are still ambient noises in some of the recordings. These background noises limit the overall performance of TTS systems. Secondly, we were not able to successfully conduct human MOS ratings, which complicates the interpretation of the results. 

Secondly, while we would like to make the collected data publicly available for replication and language documentation research, we were unable to do so this time, as we were not able to obtain the consent of the Indigenous voice donors at this moment. The primary concern is the malicious use of the data that might harm the communities. However, we will continue to work with them and aim for more open-source corpora in the long run. 

Our research currently focuses mostly on machine learning system development. To make speech technology truly beneficial to the Indigenous communities, more human-centered designs that take into consideration the community-specific cultural norms will also be needed to deploy these systems to the benefit of the Indigenous communities \cite{noe2024generalized}.

\section*{Acknowledgments}
We thank the AC and three anonymous reviewers for their insightful comments, which helped improve this article considerably.  We also thank UBC Advanced Research Computing and Digital Alliance of Canada for their computing support. This research is supported by the Mitacs Accelerate Grant awarded to CH, CQ and JZ, and the NSERC Discovery Grant awarded to JZ. This work is impossible without the contributions from the Indigenous communities.

\bibliography{custom,anthology}

\appendix

\section{Training details}
\label{app:training_details}

\begin{table*}[!ht]
    \centering
    \small

        \begin{tabular}{lllll}
        \toprule
            ~ & \textbf{Self-Attention} & \textbf{FNet} & \textbf{Mamba2} & \textbf{Hydra}  \\ \midrule
            Speaker embedding dimension & 256 & 256 & 256 & 256  \\ 
            Language embedding dimension & 192 & 192 & 192 & 192  \\ 
            Encoder hidden channels & 640 & 640 & 640 & 640  \\ 
            Encoder filter channels & 768 & 768 & 768 & 768  \\ 
            Encoder dropout rate & 0.1 & 0.1 & 0.1 & 0.1  \\ \midrule
            Decoder in channels & 160 & 160 & 160 & 160  \\ 
            Decoder out channels & 80 & 80 & 80 & 80  \\ 
            Decoder downsampling in channels & 256 & 256 & 192 & 192  \\ 
            Decoder hidden dimension & 256 & 256 & 192 & 192  \\ 
            Upsampling in channels & 256 & 256 & 192 & 192  \\ 
            Decoder hidden blocks & 2 & 2 & 2 & 2  \\ \midrule
            Optimizer type & Adam & Adam & Adam & Adam  \\ 
            Learning rate & 1.00e-06 & 1.00e-04 & 1.00e-04 & 1.00e-04  \\ 
            Scheduler & - & Cosine & Cosine & Cosine \\ \bottomrule
        \end{tabular}

    \caption{Training details, including dimensions and optimizer/scheduler information.}
    \label{table:training_details}
\end{table*}
For the purpose of replication, all training details are provided in Table \ref{table:training_details}.

\section{Benchmarking efficiency}

\begin{figure}[t]
    \includegraphics[width=0.5\textwidth]{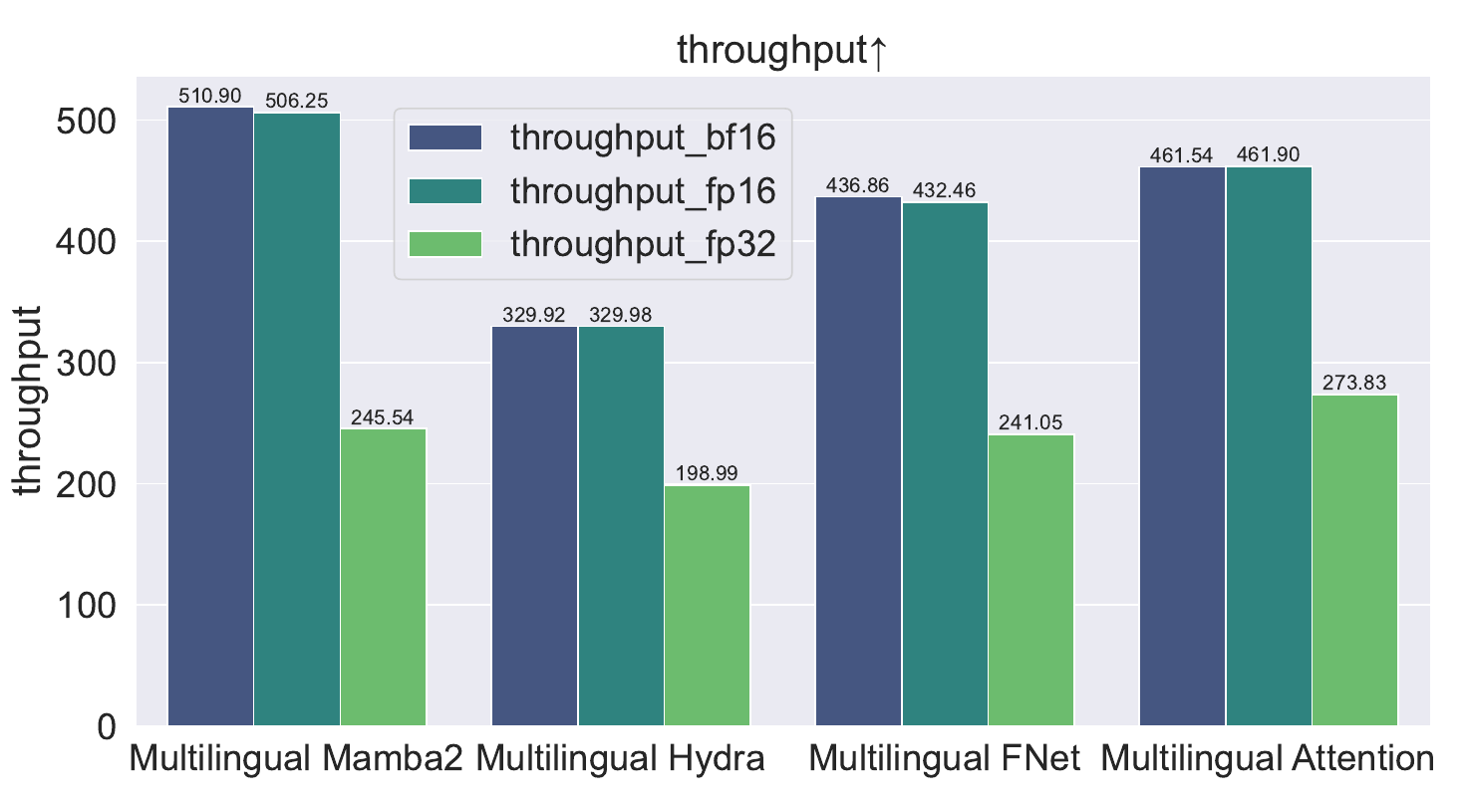}
    \caption{Throughput comparison between different models and data types. Evaluations are all performed on a single A100 GPU with a batch size of 400.}
    \label{fig:throughput}
\end{figure}

\paragraph{Throughput} We measured the throughput of each multilingual model with different data types (\texttt{bfloat16}, \texttt{float16}, and \texttt{float32}). Results are shown in Figure \ref{fig:throughput}. It can be seen that the Mamba2 model yields the highest throughput in half-precision, while Attention has the highest throughput in full-precision. FNet has slightly lower throughput than Attention, which we believe is because there is limited optimization to the kernel of the FFT algorithm. Amongst all the models, Hydra has the lowest throughput in all precisions.

\paragraph{Peak Memory Usage} We measured peak memory usage for both batched and one-by-one synthesis for all our models under using data types (\texttt{float16}, \texttt{bfloat16} and \texttt{float32}). Results are shown in Table \ref{table:memory}. It is seen that under all settings FNet is the most memory-efficient implementation as it is parameter-free. Hydra and Mamba2 have similar memory usage when performing one-by-one synthesis, but Hydra has slightly lower memory usage in batched synthesis. Attention has the highest memory usage among all models and consumed approximately twice the memory required by other implementations for batched synthesis.
\begin{table*}[!ht]
    \centering
    \small
    \begin{tabular}{ccrrrr}
    \toprule
        Data Type & Batch size & \textbf{Self-Attention} & \textbf{FNet}  & \textbf{Hydra} & \textbf{Mamba2}  \\ \midrule
        \texttt{float16} & 400 & 3.75G & 1.35G & 1.45G & 1.5G  \\
        \texttt{bfloat16} & 400 & 3.75G & 1.35G & 1.45G & 1.5G  \\
        \texttt{float32} & 400 & 4.6G & 2.3G & 2.4G & 2.5G  \\ \hline
        \texttt{float16} & 1 & 133M & 112M & 127M & 127M  \\
        \texttt{bfloat16} & 1 & 133M & 112M & 127M & 127M  \\
        \texttt{float32} & 1& 245M & 202M & 235M & 230M \\ \bottomrule
    \end{tabular}
    \caption{Peak memory usage during inference.}
    \label{table:memory}
\end{table*}

\section{Additional results}
\label{app:additional_results}
We also provide objective evaluation results using \texttt{float16} and \texttt{bfloat16} data types in Tale~\ref{table:objective_additional}. Compared to \texttt{float32}, performing in inference in \texttt{float16} and \texttt{bfloat16} data types do not bring perceptible degradation of speech quality. 

\begin{figure}[t]
    \includegraphics[width=0.5\textwidth]{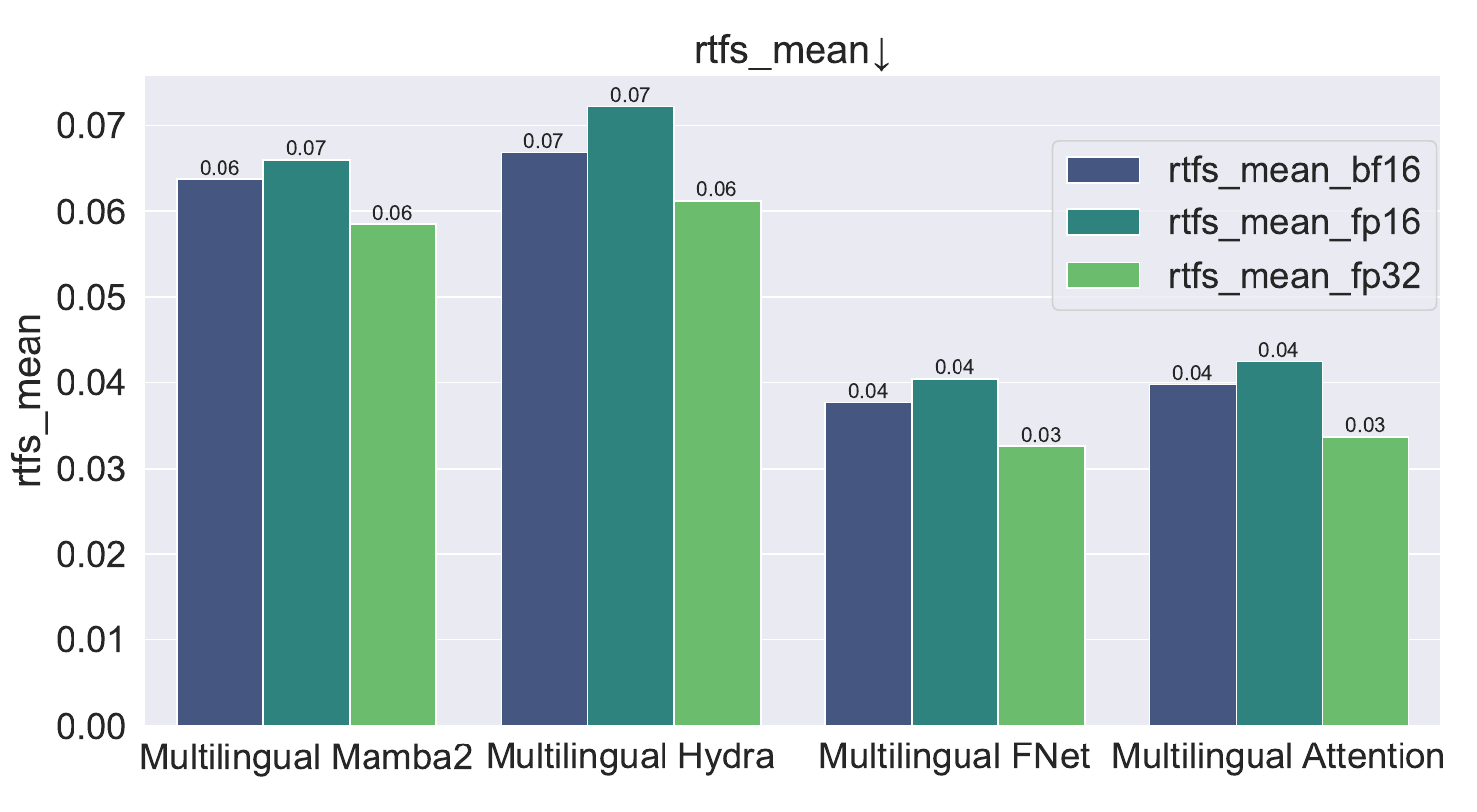}
    \caption{RTF comparison between different models and data types. Evaluations are all performed on a single A100 GPU.}
    \label{fig:rtf}
\end{figure}

\paragraph{Real Time Factor} We also measured the real time factor (RTF) of each multilingual model with different data types. Results are shown in Figure \ref{fig:rtf}. The FNet model is the fastest among all models in every setting, followed by the Attention model, Mamba2 model, and Hydra model.

\begin{table*}[p]
    \centering
    \small
    \resizebox{\textwidth}{!}{
        \begin{tabular}{lrrrrrrr}
        \toprule
            {\bf Model} & {\bf F0 RMSE↓} & {\bf LAS RMSE↓} & {\bf MCD↓} & {\bf PESQ↑} & {\bf STOI↑} & {\bf VUV F1↑} & {\bf FID↓}  \\ \midrule
            \textit{Maliseet AT float16} \\ \hline
            Monolingual Self-Attention & 84.268 & 9.064 & 19.547 & 1.217 & 0.032 & 0.655 & 0.007  \\ 
            Multilingual Mamba2 & 77.949 & 8.572 & 19.143 & 1.191 & 0.035 & 0.728 & 0.006  \\ 
            Multilingual Hydra & 79.559 & 8.396 & {\bf 18.103} & {\bf 1.453} & 0.032 & 0.735 & 0.005  \\ 
            Multilingual FNet & 76.503 & 8.945 & 19.064 & 1.259 & {\bf 0.036} & 0.723 & 0.009  \\ 
            Multilingual Attention & {\bf 75.395} & {\bf 8.045} & 18.166 & 1.344 & 0.035 & {\bf 0.745} & {\bf 0.004} \\ \midrule
            \textit{Maliseet AT bfloat16} \\ \hline
            Monolingual Self-Attention & 78.399 & 8.575 & 19.155 & 1.188 & 0.035 & 0.727 & 0.006  \\ 
            Multilingual Mamba2 & 79.160 & {\bf 8.419} & {\bf 18.101} & {\bf 1.505} & 0.030 & 0.732 & {\bf 0.005}  \\ 
            Multilingual Hydra & 77.275 & 8.981 & 19.069 & 1.250 & 0.036 & 0.719 & 0.009  \\ 
            Multilingual FNet & {\bf 73.732} & 8.082 & 18.184 & 1.346 & {\bf 0.036} & {\bf 0.741} & 0.006  \\ 
            Multilingual Attention & 80.352 & 9.021 & 19.552 & 1.243 & 0.031 & 0.657 & 0.006 \\ \midrule
            \textit{Mi'kmaq MJ float16} \\ \hline
            Monolingual Self-Attention & 139.765 & 7.820 & 22.069 & 1.139 & 0.039 & 0.641 & {\bf 0.002}  \\ 
            Multilingual Mamba2 & {\bf 138.344} & 7.369 & 21.595 & {\bf 1.248} & {\bf 0.039} & 0.644 & 0.003  \\ 
            Multilingual Hydra & 142.309 & 7.732 & 21.722 & 1.183 & 0.037 & 0.637 & 0.004  \\ 
            Multilingual FNet & 139.886 & {\bf 7.341} & {\bf 21.591} & 1.167 & 0.035 & {\bf 0.664} & 0.004  \\ 
            Multilingual Attention & 141.000 & 8.606 & 21.688 & 1.107 & 0.038 & 0.632 & 0.008 \\ \midrule
            \textit{Mi'kmaq MJ bfloat16} \\ \hline
            Monolingual Self-Attention & 139.291 & 8.616 & 21.677 & 1.110 & {\bf 0.039} & 0.631 & 0.006  \\ 
            Multilingual Mamba2 & 140.011 & 7.795 & 22.045 & 1.164 & 0.039 & 0.634 & 0.004  \\ 
            Multilingual Hydra & 138.170 & {\bf 7.078} & 21.688 & {\bf 1.196} & 0.037 & 0.653 & 0.007  \\ 
            Multilingual FNet & 142.229 & 7.693 & 21.744 & 1.169 & 0.038 & 0.630 & 0.004  \\ 
            Multilingual Attention & {\bf 138.039} & 7.310 & {\bf 21.670} & 1.161 & 0.037 & {\bf 0.663} & {\bf 0.003} \\ \midrule
            \textit{Ojibwe NJ float16} \\ \hline
            Monolingual Self-Attention & {\bf 79.762} & {\bf 6.202} & 17.565 & 1.122 & 0.034 & 0.827 & 0.009  \\ 
            Multilingual Mamba2 & 89.746 & 6.319 & 17.523 & 1.113 & 0.031 & 0.822 & {\bf 0.005}  \\ 
            Multilingual Hydra & 86.628 & 6.675 & 18.035 & 1.136 & 0.029 & 0.831 & 0.006  \\ 
            Multilingual FNet & 96.415 & 6.723 & 19.241 & 1.084 & {\bf 0.038} & 0.789 & 0.013  \\ 
            Multilingual Attention & 87.666 & 6.463 & {\bf 17.419} & {\bf 1.151} & 0.034 & {\bf 0.832} & 0.006 \\ \midrule
            \textit{Ojibwe NJ bfloat16} \\ \hline
            Monolingual Self-Attention & {\bf 80.424} & {\bf 6.215} & 17.439 & 1.116 & 0.034 & 0.830 & 0.008  \\ 
            Multilingual Mamba2 & 90.767 & 6.334 & 17.527 & {\bf 1.117} & 0.032 & 0.820 & {\bf 0.006}  \\ 
            Multilingual Hydra & 86.739 & 6.698 & 17.978 & 1.139 & 0.030 & 0.831 & 0.006  \\ 
            Multilingual FNet & 96.239 & 6.732 & 19.261 & 1.089 & {\bf 0.035} & 0.793 & 0.013  \\ 
            Multilingual Attention & 92.625 & 6.452 & {\bf 17.427} & 1.134 & 0.033 & {\bf 0.838} & 0.008 \\ \midrule
            \textit{Ojibwe JJ float16} \\ \hline
            Monolingual Self-Attention & 57.697 & 5.270 & 22.044 & 1.248 & {\bf 0.036} & 0.842 & 0.004  \\ 
            Multilingual Mamba2 & {\bf 56.191} & {\bf 4.812} & {\bf 18.348} & 1.218 & 0.032 & {\bf 0.844} & 0.004  \\ 
            Multilingual Hydra & 57.218 & 5.314 & 18.928 & {\bf 1.277} & 0.034 & 0.835 & 0.005  \\ 
            Multilingual FNet & 58.915 & 5.720 & 19.522 & 1.167 & 0.032 & 0.823 & 0.006  \\ 
            Multilingual Attention & 56.748 & 4.868 & 18.423 & 1.262 & 0.033 & 0.843 & {\bf 0.004} \\ \midrule
            \textit{Ojibwe JJ bfloat16} \\ \hline
            Ojibwe JJ & 56.987 & 5.261 & 22.065 & 1.232 & 0.038 & 0.845 & 0.005  \\ 
            Multilingual Mamba2 & 56.120 & {\bf 4.803} & 18.441 & 1.233 & 0.036 & 0.844 & 0.004  \\ 
            Multilingual Hydra & 57.142 & 5.150 & 18.860 & {\bf 1.294} & {\bf 0.038} & 0.842 & {\bf 0.003}  \\ 
            Multilingual FNet & 58.680 & 5.737 & 19.493 & 1.168 & 0.036 & 0.824 & 0.005  \\ 
            Multilingual Attention & {\bf 56.118} & 4.853 & {\bf 18.406} & 1.242 & 0.037 & {\bf 0.845} & 0.004 \\ \bottomrule
        \end{tabular}
    }
    \caption{Objective evaluation results in \texttt{float16} and \texttt{bfloat16}.}
    \label{table:objective_additional}
\end{table*}

\section{Vocoder comparison}
We compared three representative vocoders for waveform generation, namely, a time-domain vocoder \textbf{HiFi-GAN} \cite{kong2020hifi}, a frequency-domain vocoder \textbf{Vocos} \cite{siuzdak2024vocos}, and a diffusion-based vocoder \textbf{Fregrad} \cite{nguyen2024fregrad}. For HiFi-GAN, we directly used the pretrained universal HiFi-GAN\footnote{\url{https://github.com/jik876/hifi-gan}}. For both Vocos and Fregrad, we trained them on all training samples with the default parameters in their official implementation\footnote{\url{https://github.com/gemelo-ai/vocos}}\footnote{\url{https://github.com/kaistmm/fregrad}}. 
Objective results on test samples are shown in Table \ref{table:vocoders}. Since Vocos leads over other models on most objective metrics and RTF. We finally chose Vocos as our vocoder in all evaluations of TTS models. 
\label{app:vocoder}
\begin{table*}[!ht]
    \centering
    \small
    \resizebox{\textwidth}{!}{
        \begin{tabular}{lrrrrrrrr}
        \toprule
            {\bf Model} & {\bf F0 RMSE↓} & {\bf LAS RMSE↓} & {\bf MCD↓} & {\bf PESQ↑} & {\bf STOI↑} & {\bf VUV F1↑} &{\bf RTF} &{\bf Parameter} \\ \midrule
            \textit{Maliseet AT} \\ \hline
            Fregrad&10.537         & 6.431          & 11.754          & 2.449          & 0.791         & 0.916        & 0.179 & 1.78M \\
            Hifi-GAN&8.122           & 6.610          & 5.475          & 2.431          & {\bf 0.869}         & 0.907         & 0.053&13.92M \\
            Vocos& {\bf7.216 }        & {\bf 5.982}          & {\bf 5.372}          & {\bf 3.209}          & 0.835         & {\bf 0.927}         & 0.025 &7.82M \\
            \midrule
           
            \textit{Mi'kmaq MJ} \\ \hline
            Fregrad&9.239          & 6.986          & 5.011         & 2.427          & 0.908          & 0.919          & 0.177  & 1.78M \\
            Hifi-GAN& {\bf 8.432}          & 6.280        & {\bf 2.252}          & 3.092          &{\bf 0.952}         & 0.929        & 0.050 &13.92M \\
            Vocos&9.136       & {\bf 6.091}         & 3.149          & {\bf 3.391}          & 0.911        & {\bf 0.931}        & 0.026 &7.82M \\
            \midrule
            
            \textit{Ojibwe NJ} \\ \hline
            Fregrad& 8.728        & 7.437       & 13.315        & 2.501         & 0.904       & 0.949        & 0.425 & 1.78M \\
            Hifi-GAN& 8.157          & 6.947        & {\bf 6.272}        & 2.675          & {\bf 0.945}        & 0.944    & 0.062  &13.92M\\
            Vocos& {\bf 7.916 }        & {\bf 6.576}          & 6.786         & {\bf 3.070}           & 0.925        & {\bf 0.952}         & 0.038 &7.82M \\
            \midrule
            \textit{Ojibwe JJ} \\ \hline
           Fregrad&5.544         & 6.957          & 12.797        & 2.520         & 0.903        & 0.968         & 0.265 & 1.78M\\
           Hifi-GAN&6.167           & 6.536          & 5.062          & 2.516          & {\bf 0.946}        & 0.963         & 0.056 &13.92M \\
          Vocos& {\bf 5.434}         & {\bf 5.750}         & {\bf 4.389}          & {\bf 3.073}         & 0.917        & {\bf 0.974}       & 0.027&7.82M \\
            \midrule
            \bottomrule
        \end{tabular}
    }
    \caption{Objective evaluation results among vocoder models.}
    \label{table:vocoders}
\end{table*}

\section{Subjective evaluation}
\label{subjective}
Each survey included 10 generated utterances from each of the five models and 10 utterances of natural speech. This resulted in 120 total utterances for the Ojibwe survey (60 from each speaker) and 60 for the Mi'kmaq and Maliseet models. The generated utterances were created with the text from utterances that had been withheld from model training. The study was deployed through PCIbex \cite{zehr2018penncontroller} and consisted of a series of trials where a single utterance was played and participants could rate the naturalness of each sentence on a sliding scale. The data from this scale was recorded as an integer value between 1-99 with the bottom of the scale (1) labeled unnatural and the top of the scale (99) labeled natural. At the time of writing, we have only been able to recruit two participants for the evaluation of the Ojibwe language models, but plan to do more subjective evaluation in the future.

\begin{figure}
    \centering
    \includegraphics[width=\linewidth]{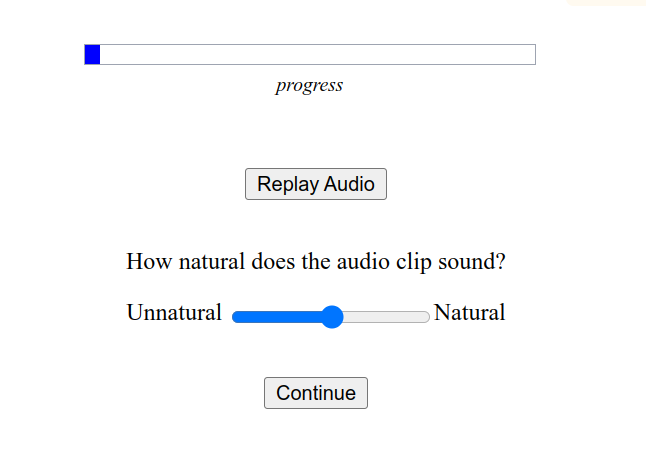}
    \caption{The MOS rating interface. }
    \label{fig:exp_interface}
\end{figure}

The participants rated speech samples by adjusting the naturalness, as shown in Fig~\ref{fig:exp_interface}. The specific instructions are given in the following textbox.

\begin{boxK}
Instructions
\begin{enumerate}
    \item A short audio clip will be played and you will be asked to rate how natural it sounds to you by toggling a sliding scale, the leftmost representing not natural at all, the rightmost representing very natural and the centre of the scale representing a neutral response
    \item Focus on the sounds of the sentence, not the meaning.
    \item There is no correct or incorrect answer, we are interested in how these audio clips sound to YOU
    \item Rate each sentence on its own, regardless of how simple or complicated it seems
\end{enumerate}

You will now move on to a practice trial where you can try rating a sample audio clip.
\end{boxK}

\end{document}